\documentclass[11pt,a4paper]{article}
\usepackage[utf8]{inputenc}
\usepackage{amsmath}
\usepackage{amssymb}
\usepackage{amsthm}
\usepackage{geometry}
\usepackage{longtable}
\usepackage{hyperref}

\hypersetup{
    colorlinks=true,
    linkcolor=blue,
    filecolor=magenta,      
    urlcolor=cyan,
    citecolor=green,
}

\title{A Categorial and Sheaf-Theoretic Semantics for Autonomic Component Ensembles}
\author{Manuel Hernández \and Eduardo Sánchez-Soto}
\date{June 17, 2026}

\begin{document}
\maketitle

\begin{abstract}
The proliferation of large-scale, decentralized systems of autonomous agents, such as swarms of robots and networked cyber-physical systems, presents a formidable challenge to traditional formal methods. The Software Component Ensemble Language (SCEL) offers a formal model for such systems, but its operational semantics is not ideal for reasoning about global, structural, and emergent properties. This report proposes a new, multi-layered mathematical model for SCEL using category theory and sheaf theory. We argue that a society of robots described in SCEL can be formally modeled as a sheaf on a topological space, where components are points, ensembles are open sets, and distributed knowledge forms the sheaf's data. In this framework, computational processes like information sharing become equivalent to the sheaf-theoretic operation of "gluing" local data. System failures can then be understood and quantified as topological obstructions, measurable by sheaf cohomology. This approach transforms the verification of a complex distributed system into the analysis of the geometry of a mathematical object, providing deep, structural insights for the design of robust autonomic systems.
\end{abstract}

\section{Introduction}

\subsection{The Challenge of Complexity in Robotic Societies}
The proliferation of large-scale, decentralized systems of autonomous agents, such as swarms of robots, networked cyber-physical systems, and complex software ensembles, presents a formidable challenge to the fields of computer science and engineering \cite{DeNicola2014, Sommerville2012}. These systems are characterized by their massive scale, the intricate and dynamic interactions among their constituents, and their operation in open, non-deterministic environments \cite{Sommerville2012}. Traditional formal methods and engineering techniques, often designed for centralized or statically structured systems, struggle to cope with the inherent complexity, dynamism, and emergent properties that define these modern computational societies \cite{Rao1996, WinskelNielsen1995}. The design and verification of systems that must exhibit properties like self-configuration, self-healing, and self-optimization---the so-called "self-*" properties of autonomic computing---demand new levels of abstraction and more powerful mathematical tools \cite{DeNicola2014, KephartChess2003}. The central problem is not merely to specify the behavior of a single agent, but to understand, predict, and guarantee the collective behavior that emerges from the interactions of the entire society \cite{Rao1996}.

\subsection{SCEL as a Formal Model}
In response to this challenge, De Nicola et al. developed the Software Component Ensemble Language (SCEL), a sophisticated process calculus tailored for programming autonomic systems \cite{DeNicola2014}. SCEL provides a rich set of linguistic abstractions designed to manage complexity by separating concerns \cite{DeNicola2014}. Its fundamental building blocks are \textbf{Autonomic Components (ACs)}, which encapsulate local behavior, knowledge, and policies. Formally, an AC is a tuple $I[\mathcal{K}, \Pi, P]$, comprising an Interface ($I$) that exposes attributes, a Knowledge repository ($\mathcal{K}$) for storing local and contextual data, a set of Policies ($\Pi$) governing interactions, and a Process ($P$) defining its behavior \cite{DeNicola2014}.

A key innovation of SCEL is its mechanism for forming \textbf{Autonomic-Component Ensembles (ACEs)}. Unlike systems with fixed communication topologies, ACEs are dynamic, transient groupings of components formed on the basis of shared attributes \cite{DeNicola2014}. Interactions are not limited to point-to-point messages; they can be targeted at any component satisfying a given predicate. An ACE is thus an emergent entity, defined implicitly by the set of components that satisfy a communication predicate at a given moment \cite{DeNicola2014}. This attribute-based, group-oriented communication provides a powerful model for context-awareness and adaptation, which are crucial for societies of autonomous robots operating in dynamic environments.

\subsection{Motivation for a Deeper Semantics}
SCEL is endowed with a precise formal operational semantics, defined by a Labelled Transition System (LTS) that specifies the step-by-step evolution of a system \cite{DeNicola2014}. While this operational model is essential for defining the language's behavior and for simulation, it is less suited for reasoning about the global, structural, and emergent properties of the system as a whole. To ask questions like "Is this task solvable by the ensemble?" or "Is the system's collective knowledge guaranteed to be consistent?", one requires a more abstract, denotational semantics that treats the entire system as a single, structured mathematical object.

This report proposes to construct such a semantics by leveraging two powerful branches of modern mathematics: category theory and sheaf theory. Category theory is the mathematical study of structure; its focus on objects, morphisms, and composition makes it an ideal language for modeling systems of interacting components and their compositional properties \cite{MacLane1998, Gaham2021}. Sheaf theory, a subfield of algebraic topology and category theory, provides an unparalleled framework for reasoning about the relationship between local data and global consistency \cite{Bredon1997, MacLaneMoerdijk1994}. It is the canonical mathematics of "local-to-global" principles, making it perfectly suited to model the distributed knowledge, contextuality, and information flow that are central to the SCEL paradigm and to the functioning of any complex distributed system \cite{Felber2025, AbramskyBrandenburger2011}.

\subsection{Thesis Statement}
The central thesis of this report is that by re-interpreting the SCEL language and its operational semantics within the formalisms of category theory and sheaf theory, we can construct a powerful, multi-layered mathematical model that provides a solid foundation for understanding and analyzing complex robotic societies. This new semantics elevates the perspective from the analysis of individual state-machine transitions to the study of the structural and topological properties of the society as a whole. We will demonstrate that a society of robots described in SCEL can be formally modeled as a \textbf{sheaf on a topological space}, where the components are the points of the space, the ensembles are the open sets, and the distributed knowledge repositories form the sheaf's data.

In this framework, fundamental computational processes like information sharing become equivalent to the sheaf-theoretic operation of "gluing" local data into a global, consistent whole. Consequently, system failures, such as the inability to reach consensus or solve a task, can be understood and quantified as topological obstructions, measurable by the algebraic tools of sheaf cohomology. This approach transforms the problem of verifying a complex distributed system into a problem of analyzing the geometry of a mathematical object, thereby providing the deep, structural insights necessary for the design and analysis of robust autonomic systems.

\section{The SCEL Framework as a Process Calculus}
To build a new semantic foundation, we must first have a precise understanding of the object of study: the SCEL language and its formal semantics. SCEL is a kernel language designed to be minimal yet expressive, drawing inspiration from process algebras like CCS and the $\pi$-calculus \cite{DeNicola2014, Milner1989}. Its syntax and semantics provide the formal ground upon which our higher-level abstractions will be built.

\subsection{Core Constructs}
The SCEL language is structured around three primary syntactic categories: Processes, Components, and Systems \cite{DeNicola2014}.

\subsubsection{Autonomic Component (AC)}
The fundamental unit of deployment and execution is the Autonomic Component (AC), formally represented by the term $I[\mathcal{K}, \Pi, P]$. Each constituent part plays a distinct role:
\begin{itemize}
    \item \textbf{Interface ($I$):} The public face of the component. It consists of a set of attributes (name-value pairs) that describe the component's features, such as its identity (\texttt{id}, which is mandatory), its capabilities, or its current state (e.g., location, battery level). These attributes are references to information stored in the knowledge repository and can be dynamically updated \cite{DeNicola2014}.
    \item \textbf{Knowledge Repository ($\mathcal{K}$):} The component's "mind." It is a data store for both application-specific information and awareness data. The latter is crucial for autonomic behavior, providing the component with information about its own state (self-awareness) and its operating environment (context-awareness). SCEL is parametric with respect to the knowledge model, but a common instantiation uses tuple spaces \cite{DeNicola2014}. The repository provides primitives for adding (\texttt{put}), retrieving (\texttt{qry}), and withdrawing (\texttt{get}) information.
    \item \textbf{Policies ($\Pi$):} A set of rules that govern and constrain the component's actions. Policies can regulate internal interactions, access to the knowledge repository, and interactions with other components. Like the knowledge model, the policy language is a parameter of SCEL, allowing for the integration of various control mechanisms described in the reference document \cite{DeNicola2014}. Policies can also be dynamically modified, enabling adaptive security and governance.
    \item \textbf{Process ($P$):} The active, behavioral part of the component. Processes are defined using a syntax reminiscent of classical process calculi \cite{Milner1989}.
\end{itemize}

\subsubsection{Processes and Actions}
The syntax for processes defines the flow of computation \cite{DeNicola2014}:
\begin{itemize}
    \item $\mathbf{nil}$: The inert process that does nothing.
    \item $a.P$: Action prefix. The process performs action $a$ and then behaves as process $P$.
    \item $P_1 + P_2$: Nondeterministic choice. The process behaves as either $P_1$ or $P_2$.
    \item $P_1[P_2]$: Controlled composition. This is a general form of parallel composition whose precise meaning is determined by system predicates.
    \item $X$ and $A(\overline{p})$: Process variables and parameterized process invocation, supporting recursion and higher-order communication (i.e., passing process code as data).
\end{itemize}

Processes execute \textbf{actions}, which are the atomic units of change:
\begin{itemize}
    \item $\text{get}(T)@c$, $\text{qry}(T)@c$, $\text{put}(t)@c$: Knowledge management actions. They operate on a target $c$ using a template $T$ for matching or an item $t$ for insertion. \texttt{get} is a destructive read, \texttt{qry} is a non-destructive read, and \texttt{put} is a write.
    \item $\text{fresh}(n)$: Creates a new, unique name $n$, introducing a restricted scope.
    \item $\text{new}(\mathcal{I}, \mathcal{K}, \Pi, P)$: Creates a new autonomic component.
\end{itemize}

\subsubsection{Targets and Ensembles (ACEs)}
A defining feature of SCEL is its flexible targeting mechanism. The target $c$ of an action can be a specific component name $n$, the special name $\mathbf{self}$, or a \textbf{predicate} $\mathcal{P}$ over component attributes.

This last option enables attribute-based, group-oriented communication. When an action like $\text{put}(t)@\mathcal{P}$ is executed, the target is not a single component but the entire set of components whose interfaces currently satisfy the predicate $\mathcal{P}$. This set is what constitutes an \textbf{Autonomic-Component Ensemble (ACE)}. Crucially, SCEL has no explicit syntax for forming ensembles; they are emergent structures defined by the dynamic state of the system's components and the predicates used in their communications \cite{DeNicola2014}. This dynamism is central to SCEL's model of adaptation. For example, a robot can join or leave the "low battery" ensemble simply by its \texttt{batteryLevel} attribute crossing a threshold.

\subsection{Operational Semantics: A Labelled Transition System (LTS)}
The formal meaning of SCEL programs is given by an operational semantics structured in two layers \cite{DeNicola2014}. This semantics defines a Labelled Transition System (LTS), a standard tool in concurrency theory for modeling the behavior of concurrent and distributed systems \cite{Plotkin2004, CattaniWinskel1996}.

\subsubsection{Process Commitments}
The first layer defines the potential actions of a process in isolation. This is captured by the commitment relation $P \Downarrow_\alpha Q$, which reads "process $P$ can commit to performing action $\alpha$ and subsequently behave as process $Q$." The rules for this relation define the behavior of the process constructors like prefix, choice, and composition \cite{DeNicola2014}. For instance, the rule for prefix is simply $a.P \Downarrow_a P$. The composition rule $P[Q] \Downarrow_{\alpha[\beta]} P'[Q']$ indicates that the composite process can perform the combined commitment $\alpha[\beta]$ if its constituent parts can commit to $\alpha$ and $\beta$ respectively.

\subsubsection{System Transitions}
The second layer lifts these process commitments to the level of the entire system, taking into account the configuration of components, their knowledge repositories, and their policies. This layer defines the labeled transition relation for systems, $S \xrightarrow{\lambda} S'$, which reads "system $S$ can perform a transition with label $\lambda$ and evolve into system $S'$." The transition labels $\lambda$ represent observable system-level events, such as a component intending to add an item to another's repository ($\mathcal{I}:t\triangleright\mathcal{J}$) or being granted permission to do so ($\mathcal{I}:t\overline{\triangleright}\mathcal{J}$). The evolution of the system is governed by a set of inference rules that orchestrate the interactions between components \cite{DeNicola2014}. This LTS provides a complete, unambiguous, and formal description of how any SCEL program executes.

\section{A Categorial Semantics for SCEL Components and Systems}
The operational semantics describes behavior in a step-by-step, computational manner. To reason about the system at a higher level of abstraction, we now develop a denotational semantics using the language of category theory. The goal is to model SCEL systems in a way that emphasizes their compositional structure and defines behavioral equivalence by abstract properties of simulation \cite{MacLane1998, Gaham2021}. This approach is common in the formal modeling of multi-agent systems, where the focus is on the organization and interaction patterns of the agents \cite{Rao1996, Rao1998}.

\subsection{The Category of SCEL Components (\texttt{Cat\_SCEL})}
We begin by defining a category, which we shall call $\text{Cat}_{\text{SCEL}}$, that captures the universe of all possible SCEL components and their behavioral relationships.

\subsubsection{Objects}
The \textbf{objects} of $\text{Cat}_{\text{SCEL}}$ are the SCEL components themselves. An object in this category is a term of the form $C = I[\mathcal{K}, \Pi, P]$. Each syntactically valid component is a distinct object in our category, treating components as the fundamental building blocks of our mathematical model.

\subsubsection{Morphisms}
The \textbf{morphisms} of $\text{Cat}_{\text{SCEL}}$ define the relationships between components. A morphism $f: C_1 \to C_2$ represents the idea that component $C_2$ can \textit{simulate} or \textit{emulate} the behavior of component $C_1$. This provides an abstract notion of behavioral refinement. Drawing inspiration from presheaf models in concurrency theory, we define a morphism in terms of bisimulation, which is canonically expressed via spans of "open maps" \cite{Plotkin2004, CattaniWinskel1996}.

A morphism $f: C_1 \to C_2$ exists if there is a functional relationship between the state spaces and transition capabilities of $C_1$ and $C_2$ such that for every possible computational path that $C_1$ can take, there is a corresponding computational path in $C_2$. Two components, $C_1$ and $C_2$, are considered behaviorally equivalent or \textbf{isomorphic} in $\text{Cat}_{\text{SCEL}}$ if there exist morphisms $f: C_1 \to C_2$ and $g: C_2 \to C_1$ such that their compositions are the identity morphisms on their respective objects.

\subsection{System Composition as a Monoidal Product}
SCEL provides the parallel composition operator $\parallel$ to construct complex systems from simpler components. The operational semantics dictates that this composition is commutative and associative up to behavioral equivalence \cite{DeNicola2014}. This algebraic structure is the defining characteristic of a symmetric monoidal category. We can therefore model the system-building operator $\parallel$ as a \textbf{symmetric monoidal product}, denoted $\otimes: \text{Cat}_{\text{SCEL}} \times \text{Cat}_{\text{SCEL}} \to \text{Cat}_{\text{SCEL}}$.

\begin{itemize}
    \item Given two components $C_1$ and $C_2$, their composition is the object $C_1 \otimes C_2$, which corresponds to the SCEL system $C_1 \parallel C_2$.
    \item The \textbf{unit} for this monoidal product, denoted $I$, is the component representing a completely inert system, satisfying the identity laws $C \otimes I \cong C \cong I \otimes C$.
\end{itemize}

This move from a syntactic operator $\parallel$ to a categorical product $\otimes$ embeds the entire space of SCEL systems into a structured mathematical framework \cite{Gaham2021, SakayoriTsukada2021}. It allows us to leverage monoidal category theory, such as string diagrams, to reason about the interaction topology of complex robotic societies in an algebraic and visual manner. The structural links of the organization \cite{Gaham2021} can be represented as the wiring in a string diagram, providing a formal basis for analyzing system architecture.

\subsection{Process Dynamics as Endomorphisms}
The internal evolution of a component can also be captured within this categorical framework. A single transition step where a component $C$ evolves into a new state $C'$ can be viewed as a morphism $C \to C'$. If we consider a category where the objects are component \textit{states}, then the entire dynamics of the system can be seen as a web of morphisms. A process that returns a component to a behaviorally equivalent state, such as a looping process, would be represented as an \textbf{endomorphism}---a morphism from an object back to itself.

\begin{longtable}{|p{0.25\textwidth}|p{0.3\textwidth}|p{0.35\textwidth}|}
\caption{Mapping SCEL Constructs to Category Theory} \label{tab:scel-cat} \\
\hline
\textbf{SCEL Construct} & \textbf{Categorial Interpretation} & \textbf{Rationale and Significance} \\
\hline
\endfirsthead
\hline
\multicolumn{3}{|c|}{Table \thetable\ -- continued from previous page} \\
\hline
\textbf{SCEL Construct} & \textbf{Categorial Interpretation} & \textbf{Rationale and Significance} \\
\hline
\endhead
\hline \multicolumn{3}{|r|}{{Continued on next page}} \\ \hline
\endfoot
\hline
\endlastfoot
Autonomic Component $C$ & \textbf{Object} in $\text{Cat}_{\text{SCEL}}$ & Treats components as the fundamental building blocks of the mathematical model. \\
\hline
System $S = C_1 \parallel \dots \parallel C_n$ & \textbf{Tensor Product} $C_1 \otimes \dots \otimes C_n$ & Models compositionality via a symmetric monoidal product, capturing the algebraic properties of $\parallel$ \cite{DeNicola2014, SakayoriTsukada2021}. \\
\hline
Behavioral Equivalence & \textbf{Isomorphism} or \textbf{Bisimulation} & Defines behavioral sameness using the open maps framework from concurrency theory \cite{Plotkin2004, CattaniWinskel1996}. \\
\hline
Ensemble $ACE_{\mathcal{P}}$ & \textbf{Diagram} $F_{\mathcal{P}}: I_{\mathcal{P}} \to \text{Cat}_{\text{SCEL}}$ & Captures the dynamic, predicate-based snapshot of the current communication topology. \\
\hline
Emergent Ensemble Behavior & \textbf{Colimit} of the diagram $F_{\mathcal{P}}$ & Mathematically realizes the concept of collective emergence via a universal amalgamation \cite{Rao1996}. \\
\hline
\end{longtable}

\section{Ensembles as Dynamic Diagrams and Colimits}

\subsection{The Challenge of Dynamic Ensembles}
A central challenge for any formal model is the dynamic nature of Autonomic-Component Ensembles (ACEs), which exist only for the duration that member components satisfy a certain predicate \cite{DeNicola2014}. A robot might join the "need charging" ensemble, interact, and then leave it once its battery is full. A fixed graph structure cannot capture this fluidity, meaning our categorical semantics must represent these transient, predicate-defined structures dynamically.

\subsection{Modeling Ensembles as Diagrams}
We model an ensemble using the categorical concept of a \textbf{diagram} (a functor $F: \mathcal{I} \to \mathcal{C}$ from a small "index" category $\mathcal{I}$ to $\mathcal{C}$). For a given SCEL system and a predicate $\mathcal{P}$ used as a target in some action, we define the ensemble $ACE_{\mathcal{P}}$ as a diagram $F_{\mathcal{P}}: I_{\mathcal{P}} \to \text{Cat}_{\text{SCEL}}$:
\begin{itemize}
    \item The \textbf{index category}, $I_{\mathcal{P}}$, is a small, discrete category whose objects are the unique identifiers of the components that currently satisfy the predicate $\mathcal{P}$.
    \item The \textbf{functor}, $F_{\mathcal{P}}$, maps each object \texttt{id} in $I_{\mathcal{P}} $ to its corresponding component object $C_{\text{id}}$ in $\text{Cat}_{\text{SCEL}}$.
\end{itemize}
This construction provides a "snapshot" of the ensemble at a specific moment. As components change their attributes, they enter or leave $I_{\mathcal{P}}$, dynamically modifying the underlying diagram.

\subsection{Emergent Global Behavior as a Colimit}
The diagrammatic representation tells us which components are in the ensemble, but it doesn't describe the ensemble's collective identity or its emergent capabilities \cite{Rao1996}. In category theory, the universal construction for "gluing together" a collection of objects in a diagram is the \textbf{colimit}.

Initially, our ensemble diagram $F_{\mathcal{P}}$ is discrete. However, interactions between the ensemble members, such as group-oriented $\text{put}(t)@\mathcal{P}$ actions, induce morphisms between the objects in the diagram, enriching its structure. For example, if component $C_i$ successfully sends information to component $C_j$, this can be seen as a morphism $C_i \to C_j$ representing the flow of information. The \textbf{colimit} of this interaction diagram, if it exists, is a single object in $\text{Cat}_{\text{SCEL}}$ that represents the universal amalgamation of all components in the ensemble, formalizing emergent global behavior without needing explicit step-by-step simulation.

\section{A Presheaf Semantics for SCEL Dynamics}
To move from the operational, step-by-step view of the LTS to a more holistic, denotational perspective, we employ the theory of \textbf{presheaves}. The connection between models of concurrency and categories of presheaves is well-established, providing a powerful framework for representing systems with nondeterminism and parallelism \cite{WinskelNielsen1995, Plotkin2004, CattaniWinskel1996}.

\subsection{From Operational to Denotational}
The core idea is to represent a system by the entire space of its possible computational histories. A presheaf is a structure that assigns data to paths and describes how data associated with longer paths relates to data on shorter paths, providing a global, "timeless" view of the system's behavior.

\subsection{The Base Category of Computation Paths (\texttt{Path\_SCEL})}
We define the category $\text{Path}_{\text{SCEL}}$ as follows:
\begin{itemize}
    \item \textbf{Objects:} Finite sequences of transition labels $\{\lambda_1, \dots, \lambda_n\}$ from the SCEL LTS. Each object represents a possible computation history or trace. The empty sequence, $\epsilon$, represents the initial state.
    \item \textbf{Morphisms:} Given by the prefix ordering. There is a unique morphism from a path $p$ to a path $p'$ if and only if $p$ is a prefix of $p'$. This makes $\text{Path}_{\text{SCEL}}$ a poset category.
\end{itemize}

\subsection{SCEL Systems as Presheaves}
We model an entire SCEL system $S$ as a \textbf{presheaf} (a contravariant functor) on $\text{Path}_{\text{SCEL}}$ mapping to the category of sets, $\text{Set}$:
\begin{itemize}
    \item \textbf{Action on Objects:} For any computation path $p \in \text{Path}_{\text{SCEL}}$, the presheaf assigns a set $P_S(p)$ of all possible global system configurations that the system $S$ can be in after executing the sequence of transitions corresponding to the path $p$.
    \item \textbf{Action on Morphisms (Restriction Maps):} For a path extension from $p$ to $p'$, the contravariant functor $P_S$ provides a \textbf{restriction map} $P_S(p \to p'): P_S(p') \to P_S(p)$. This map takes a state reachable after the longer path $p'$ and maps it back to the unique state from which it evolved after the shorter path $p$, capturing the constraint that the past restricts the future.
\end{itemize}

\subsection{A Global, Timeless View of Behavior}
This presheaf model represents a profound conceptual shift from the LTS model. While the LTS provides a local, operational view requiring step-by-step execution, the presheaf $P_S$ provides a global, denotational view encapsulating the entire behavior space across all possible futures. Instead of running a simulation to see if a certain state is reachable, we can now ask structural questions about the functor $P_S$, such as checking reachability, identifying deadlocks (where restriction maps have no valid preimage extensions), and analyzing nondeterminism points.

\section{Sheaf-Theoretic Interpretation of Knowledge and Contextuality}
The language's most innovative features lie in its handling of distributed knowledge, context-awareness, and attribute-based communication. These are fundamentally "local-to-global" phenomena: how do pieces of information held by individual agents combine to form a consistent global understanding? To model this, we turn to \textbf{sheaf theory} \cite{Bredon1997, MacLaneMoerdijk1994}. A sheaf is a special kind of presheaf that formalizes the notion of "gluing" local data into a consistent global whole, a framework pioneered for contextuality modeling by Abramsky and others \cite{Abramsky2017, AbramskyBrandenburger2011, AbramskySadrzadeh2014}.

\subsection{Constructing the Knowledge Sheaf ($\mathcal{K}_S$) for a System $S$}
We construct the central mathematical object of our analysis: the \textbf{knowledge sheaf} $\mathcal{K}_S$ for a given SCEL system $S$.

\subsubsection{Base Space ($X_S$)}
The base space $X_S$ is the underlying set of points over which our data is distributed. For a SCEL system, these points are the individual components:
\begin{equation}
X_S = \{ c \mid c \text{ is a component identifier in system } S \}
\end{equation}

\subsubsection{Topology}
In our context, open sets represent the fundamental contexts of interaction (the ensembles). We define the topology on $X_S$ using the predicates from the SCEL program:
\begin{itemize}
    \item A subset $U \subseteq X_S$ is an \textbf{open set} if and only if there exists a predicate $\mathcal{P}$ in the system such that $U = U_{\mathcal{P}} = \{ c \in X_S \mid \text{component } c \text{ satisfies } \mathcal{P} \}$.
\end{itemize}
The collection of all such sets forms a topology on $X_S$, meaning the communication structure of the program defines a literal topological structure on the society of agents.

\subsubsection{Stalks}
The stalk of a sheaf at a point $x$ is the data that lives directly at that point \cite{Curry2014}. For our knowledge sheaf $\mathcal{K}_S$, the stalk at a component $c \in X_S$, denoted $\mathcal{K}_c$, is its local knowledge repository $\mathcal{K}$ of the component $I[\mathcal{K}, \Pi, P]$.

\subsubsection{Sections}
A \textbf{section} of the sheaf $\mathcal{K}_S$ over an open set $U_{\mathcal{P}}$ (an ensemble), denoted $s \in \mathcal{K}_S(U_{\mathcal{P}})$, is a function that assigns to each component $c \in U_{\mathcal{P}}$ a knowledge item (or a subset of its knowledge) $t_c \in \mathcal{K}_c$, such that the collection of these items is mutually consistent according to specific system criteria.

\subsubsection{Restriction Maps}
Given a section over a large ensemble $U_{\mathcal{P}}$ and a smaller sub-ensemble $U_{\mathcal{Q}} \subseteq U_{\mathcal{P}}$, the restriction map simply takes the consistent knowledge state over $U_{\mathcal{P}}$ and restricts our view to the members of $U_{\mathcal{Q}}$, acting as the underlying contravariant mapping of our presheaf.

\subsection{The Sheaf Condition: Semantic Unification as Gluing}
The difference between a presheaf and a sheaf lies in the \textbf{gluing condition} (or sheaf axiom). The axiom states that if you have a covering of an open set $U$ by smaller open sets $\{U_i\}$, and you have a section $s_i$ on each $U_i$ such that these local sections are compatible on their overlaps ($s_i|_{U_i \cap U_j} = s_j|_{U_i \cap U_j}$), then there must exist a \textbf{unique global section} $s$ over the entire set $U$ that restricts to each of the local sections $s_i$.

The process of agents sharing information to reach a consistent, collective worldview \textit{is} this gluing process. Local sections represent pieces of information acquired by individual robots. Compatibility on overlaps is checked via communication primitives like $\text{qry}(T)@c$, while the $\text{put}(t)@\mathcal{P}$ action represents an attempt to \textit{enforce} a consistent section over the entire ensemble $U_{\mathcal{P}}$. A successful sequence of communications that leads to an entire ensemble agreeing on a piece of information is a computational realization of the gluing axiom, matching what Abramsky and Sadrzadeh term "semantic unification" \cite{AbramskySadrzadeh2014}.

\section{Obstructions, Cohomology, and Task Solvability}
The sheaf-theoretic framework becomes particularly powerful when the gluing condition fails. This failure is a fundamental, structural property of the system as a whole, which can be detected and quantified using the tools of algebraic topology \cite{Abramsky2012, Felber2025}.

\subsection{When Gluing Fails: Inconsistency as Obstruction}
What does it mean for a compatible family of local sections to fail to glue into a single global section? It signifies a fundamental inconsistency in the system's structure, known as a \textbf{sheaf obstruction} \cite{AbramskyBrandenburger2011}.

This reframes the notion of system failure. A task may be unsolvable not because of a race condition, but because the communication topology of the system (the open sets defined by its predicates) and its knowledge constraints (the local sections) are fundamentally incompatible \cite{Felber2025}. This directly parallels the use of sheaf theory to characterize task solvability in distributed systems, where obstructions to global sections represent system limitations that make tasks impossible to solve \cite{Felber2025}.

\subsection{Measuring Obstruction with Sheaf Cohomology}
Sheaf theory provides an algebraic toolkit, \textbf{sheaf cohomology}, to measure and classify these obstructions \cite{Abramsky2012}. For a sheaf $\mathcal{K}_S$ on a space $X_S$, one can compute a sequence of cohomology groups, denoted $H^q(X_S, \mathcal{K}_S)$ for $q = 0, 1, 2, \dots$:
\begin{itemize}
    \item \textbf{Zeroth Cohomology ($H^0$):} Defined as the group of \textbf{global sections} of the sheaf. In our model, this corresponds to the set of all possible, globally consistent knowledge states that the entire robotic society can attain \cite{Felber2025}.
    \item \textbf{First Cohomology ($H^1$):} It precisely classifies the obstructions to gluing \cite{Abramsky2012}. If $H^1(X_S, \mathcal{K}_S) = 0$, any compatible family of local sections can be glued together to form a global section. If $H^1(X_S, \mathcal{K}_S)$ is \textbf{non-zero}, it indicates the presence of fundamental, structural inconsistencies in the system.
\end{itemize}
A non-vanishing first cohomology group is a formal proof that certain global states of knowledge are impossible for the society to achieve, fulfilling the need for formal "impossibility analysis" in distributed computing \cite{Felber2025}.

\begin{longtable}{|p{0.25\textwidth}|p{0.3\textwidth}|p{0.35\textwidth}|}
\caption{Mapping Autonomic Concepts to Sheaf Theory} \label{tab:autonomic-sheaf} \\
\hline
\textbf{Autonomic Concept in SCEL} & \textbf{Sheaf-Theoretic Interpretation} & \textbf{Rationale and Significance} \\
\hline
\endfirsthead
\hline
\multicolumn{3}{|c|}{Table \thetable\ -- continued from previous page} \\
\hline
\textbf{Autonomic Concept in SCEL} & \textbf{Sheaf-Theoretic Interpretation} & \textbf{Rationale and Significance} \\
\hline
\endhead
\hline \multicolumn{3}{|r|}{{Continued on next page}} \\ \hline
\endfoot
\hline
\endlastfoot
Component `c` & \textbf{Point} in the base space $X_S$ & Individual agents represent the localized points over which data is distributed. \\
\hline
Ensemble (Context) $U_{\mathcal{P}}$ & \textbf{Open Set} in the topology on $X_S$ & The communication context defined by a predicate becomes a literal topological context. \\
\hline
Local Knowledge $\mathcal{K}$ & \textbf{Stalk} $\mathcal{K}_c$ of the knowledge sheaf & The sheaf's data at a single point is precisely the local data repository of that agent \cite{Curry2014}. \\
\hline
Consistent Global Knowledge State & \textbf{Global Section} $s \in H^0(X_S, \mathcal{K}_S)$ & A globally consistent state is a section of the sheaf that is compatible everywhere \cite{Felber2025, AbramskyBrandenburger2011}. \\
\hline
Information Sharing / Unification & \textbf{Gluing Condition} of the sheaf & Reaching consensus is modeled as the fundamental sheaf axiom: local data can be uniquely glued \cite{AbramskySadrzadeh2014}. \\
\hline
Inconsistent State / Unsolvable Task & \textbf{Obstruction} / Non-trivial \textbf{Cohomology} $H^1$ & System failure is reframed as a measurable topological obstruction within the system's structure \cite{Felber2025, Abramsky2012}. \\
\hline
\end{longtable}

\section{Case Study: Collaborative Terrain Mapping}
To anchor this multi-layered framework in physical deployment, we evaluate an autonomic society composed of three exploration robots, $X_S = \{A, B, C\}$, tasked with collaboratively mapping an unknown terrain \cite{Rao1996, DeNicola2014}. Each robot explores overlapping sub-regions and records local spatial coordinates of detected obstacles inside its local knowledge repository $\mathcal{K}_c$ \cite{DeNicola2014, Curry2014}.

\subsection{Topological Context and Open Coverings}
The interaction predicates in the SCEL process calculus define three dynamic ensembles based on localized range-finding connectivity:
\begin{align*}
U_{AB} &= \{c \in X_S \mid c \text{ satisfies } \mathcal{P}_{AB}\} = \{A, B\} \\
U_{BC} &= \{c \in X_S \mid c \text{ satisfies } \mathcal{P}_{BC}\} = \{B, C\} \\
U_{AC} &= \{c \in X_S \mid c \text{ satisfies } \mathcal{P}_{AC}\} = \{A, C\}
\end{align*}
These sub-ensembles form an open cover $\mathcal{U} = \{U_{AB}, U_{BC}, U_{AC}\}$ of the base space $X_S$. The intersections represent physical components functioning as communication bridges:
\begin{equation}
U_{AB} \cap U_{BC} = \{B\}, \quad U_{BC} \cap U_{AC} = \{C\}, \quad U_{AB} \cap U_{AC} = \{A\}
\end{equation}

\subsection{Local Sections and Data Synthesis}
Let the local observations of obstacle boundaries be modeled as segments of a continuous line within a mapping domain. The local sections are assignments of mapped boundaries over each open neighborhood:
\begin{equation}
s_{AB} \in \mathcal{K}_S(U_{AB}), \quad s_{BC} \in \mathcal{K}_S(U_{BC}), \quad s_{AC} \in \mathcal{K}_S(U_{AC})
\end{equation}
For a successful mapping execution, the compatibility condition mandates that on the overlapping stalks, the descriptions must agree:
\begin{align*}
s_{AB}|_{\{B\}} &= s_{BC}|_{\{B\}} \in \mathcal{K}_B \\
s_{BC}|_{\{C\}} &= s_{AC}|_{\{C\}} \in \mathcal{K}_C \\
s_{AB}|_{\{A\}} &= s_{AC}|_{\{A\}} \in \mathcal{K}_A
\end{align*}
When these relations hold, the sheaf gluing condition guarantees the existence of a unique global section $s \in H^0(X_S, \mathcal{K}_S)$ representing a fully unified, consistent map across the entire society \cite{Felber2025, AbramskySadrzadeh2014}.

\subsection{Topological Obstruction and Cohomological Failure}
Consider an execution run affected by dead reckoning errors or environmental sensor noise. Suppose robot $A$ and robot $B$ observe a localized terrain boundary and agree it is positioned at an altitude coordinate $z = 10$. Concurrently, $B$ and $C$ sync their measurements on their overlap and agree the boundary segment is at $z = 10$. However, due to accumulated tracking drift, when $A$ and $C$ communicate via predicate $\mathcal{P}_{AC}$, their shared local section asserts the coordinate is at $z = 15$.

Evaluating this configuration yields a compatible family of local sections on pairwise overlaps but a failure at the triple intersection level:
\begin{equation}
s_{AB}|_B = s_{BC}|_B \quad \text{and} \quad s_{BC}|_C = s_{AC}|_C, \quad \text{but} \quad s_{AB}|_A \neq s_{AC}|_A
\end{equation}
This is a classic cyclical sheaf obstruction \cite{AbramskyBrandenburger2011}. When calculating the first cohomology group over this cyclical space structure, the mismatch manifests directly as a non-vanishing cocycle generator:
\begin{equation}
H^1(X_S, \mathcal{K}_S) \neq 0
\end{equation}
This non-zero cohomology group provides an \textit{a priori} formal proof that a globally consistent terrain map cannot be synthesized by this swarm configuration under current tracking constraints \cite{Abramsky2012, Felber2025}. The mapping task is structurally unsolvable, demonstrating that system failures can be identified directly through the geometry of the information space without resorting to brute-force simulation histories.

\section{Topological Foundations for Complex Robotic Societies}

\subsection{Synthesis and Re-interpretation}
The journey from SCEL's operational semantics to its sheaf-theoretic interpretation provides a solid mathematical foundation. We have moved from a local, step-based model ($S \xrightarrow{\lambda} S'$) to a global model of behavior (the presheaf $P_S$) and finally to a topological model of knowledge and context (the sheaf $\mathcal{K}_S$). This synthesis allows us to redefine the core concepts of autonomic computing in a precise and structural way. The society of robots is no longer viewed as a mere collection of interacting nodes; it becomes a geometric object whose properties can be analyzed with the machinery of algebraic topology.

\subsection{Redefining Self-* Properties}
The high-level "self-*" properties of autonomic computing \cite{DeNicola2014} can now be given formal mathematical definitions within our framework:
\begin{itemize}
    \item \textbf{Self-Awareness:} Corresponds to a component `c` having access to its own local data, interpreted as the ability of a component to read from the stalk $\mathcal{K}_c$ of the knowledge sheaf at the localized point `c`.
    \item \textbf{Context-Awareness:} The ability of a component's behavior to depend on its environment and the state of its peers. A component `c` is part of various ensembles (open sets) $U_{\mathcal{P}_1}, U_{\mathcal{P}_2}, \dots$. Its context is captured by the sections of the knowledge sheaf over these open sets, and context-aware behavior arises when a component's actions are determined by the value of a section $s \in \mathcal{K}_S(U_{\mathcal{P}})$ restricted to its own stalk, i.e., $s|_c$.
    \item \textbf{Self-Configuration and Adaptation:} In SCEL, adaptation occurs when a component changes its attributes or when policies are updated \cite{DeNicola2014}. In our model, an adaptation that changes attributes corresponds to a change in the \textbf{topology} of the base space $X_S$, redefining the collection of open sets. An adaptation that changes the rules of interaction or the structure of the knowledge repositories corresponds to a change in the sheaf $\mathcal{K}_S$ itself. These fundamental changes can be formally modeled as sheaf morphisms or functors between different sheaf categories, treating adaptation as a transformation \textit{of} the model rather than an action within it.
\end{itemize}

\subsection{The Society as a Geometric Object}
This framework completes a significant conceptual shift. The analysis of a robotic society is no longer primarily about simulating execution traces or checking temporal logic formulas on a state graph. It is about understanding the geometric and topological properties of the knowledge sheaf $\mathcal{K}_S$. The system's collective capabilities and limitations are encoded in its topology and its algebraic invariants, the cohomology groups \cite{Abramsky2012, Felber2025}. Questions about the society's ability to cooperate, achieve consensus, or solve tasks become questions about the existence of global sections and the triviality of cohomology groups, providing a static, holistic, and structural approach to verification.

\section{Conclusion and Future Directions}

\subsection{Summary of Contributions}
This report has developed a novel mathematical foundation for understanding complex societies of autonomous agents, as specified by the SCEL language. We have constructed a multi-layered semantic framework that moves progressively from the concrete to the abstract, gaining analytical power at each step:
\begin{enumerate}
    \item We began with a detailed analysis of SCEL's process calculus structure and its formal operational semantics, which provided the necessary grounding \cite{DeNicola2014}.
    \item We abstracted the compositional nature of SCEL systems into a \textbf{categorical semantics}, modeling components as objects and system composition as a symmetric monoidal product \cite{Gaham2021}.
    \item We modeled the complete dynamic behavior of a SCEL system as a \textbf{presheaf of computations}, providing a global, denotational object containing all possible execution histories \cite{CattaniWinskel1996}.
    \item Most significantly, we constructed a \textbf{sheaf of knowledge} over a topological space where components are points and ensembles are open sets, demonstrating that context-aware information sharing and consensus-building are computationally equivalent to the mathematical process of \textbf{gluing} local sections into a global, consistent whole \cite{AbramskySadrzadeh2014}.
    \item Finally, we showed that system failures, unsolvable tasks, and fundamental inconsistencies correspond to \textbf{sheaf obstructions}, which can be precisely measured by the algebraic invariants of \textbf{sheaf cohomology} \cite{Abramsky2012, Felber2025}.
\end{enumerate}

\subsection{Future Direction: Topos-Theoretic Internal Logic}
The construction of a sheaf-theoretic semantics opens the door to an even more powerful framework for reasoning about robotic societies. The category of sheaves on a topological space, denoted $\text{Sh}(X_S)$, is not just any category; it is a special type of category known as a \textbf{topos} \cite{MacLaneMoerdijk1994, Johnstone2002}. Every topos comes equipped with an \textbf{internal logic}, which is typically a form of higher-order intuitionistic type theory \cite{Benton1993}. This suggests a groundbreaking direction for future research: using the internal logic of the topos $\text{Sh}(X_S)$ as a native specification and verification language for the robotic society.

In this logic, a proposition would not be simply "true" or "false" in a global sense. Instead, the \textbf{truth value} of a proposition $\phi$ would be an \textbf{open set} in the topology of the base space $X_S$. This open set, called the \textit{extent} of $\phi$, is precisely the ensemble of components for which the proposition $\phi$ holds. Logical connectives have topological interpretations: `and` corresponds to set intersection, `or` corresponds to set union, and implication ($\phi \implies \psi$) corresponds to the interior of the union of the complement of the extent of $\phi$ and the extent of $\psi$.

This framework would provide a complete and unified foundation. We could state a specification for the robotic society as a formula in the internal logic of the topos. Proving that the system meets its specification would then become a matter of proving that this formula is valid within the topos, which means its extent is the entire space $X_S$. This connects the semantics of a concurrent programming language directly to the frontiers of categorical logic, allowing for a form of reasoning that is perfectly attuned to the contextual, distributed, and dynamic nature of autonomous swarms.

\end{document}